\documentclass[runningheads]{llncs}
\usepackage[T1]{fontenc}
\usepackage{placeins}
\usepackage{multirow}
\usepackage{color}
\usepackage{graphicx}
\usepackage{subfigure}
\usepackage{amsmath,amssymb,amsfonts}
\usepackage{soul}
\usepackage{color}
\usepackage{ulem}
\usepackage{hyperref}
\hypersetup{
    colorlinks=true,
    linkcolor=blue,
    filecolor=magenta,      
    urlcolor=cyan,
    pdftitle={Overleaf Example},
    pdfpagemode=FullScreen,
    }
\begin{document}
\title{Learning Transferable Object-Centric Diffeomorphic Transformations for Data Augmentation in Medical Image Segmentation}
\author{Nilesh Kumar\inst{1} \and
Prashnna K. Gyawali\inst{2} \and Sandesh Ghimire\inst{3} \and Linwei Wang\inst{1} 
} %index{Kumar, Nilesh} %index{Gyawali, Prashnna} %index{Ghimire, Sandesh} %index{Wang, Linwei}
\institute{Rochester Institute of Technology, NY, USA\\
\email{nk4856@rit.edu}
\and{West Virginia University}
\and{Qualcomm Inc}}

\maketitle              % typeset the header of the contribution

\begin{abstract}
Obtaining labelled data in medical image segmentation is challenging due to the need for pixel-level annotations by experts. 
Recent works have shown that 
augmenting the object of interest with deformable transformations
can help mitigate this challenge.
However, these transformations have been learned globally for the image, limiting their transferability across datasets or applicability in 
problems where image alignment is difficult.
While object-centric augmentations provide a great opportunity to overcome these issues, existing works
 are only focused on position and random transformations without considering shape variations of the objects. 
 To this end, we propose a novel object-centric data augmentation model that is able to learn the shape variations for the objects of interest and augment the object in place without modifying the rest of the image. We demonstrated its effectiveness in improving kidney tumour segmentation when leveraging shape variations learned both from within the same dataset and transferred from external datasets.    

\end{abstract}

\section{Introduction}

A must-have ingredient for training a deep neural network (DNN) is a large number of labelled data that is not always available in real-world applications. %including biomedical images. 
This challenge of data annotation becomes even worse for medical image segmentation tasks that require pixel-level annotation by experts. 
Data augmentation (DA) is a recognized approach to tackle this challenge. 
Common DA strategies create new samples by using predefined transformations such as rotation, translation, 
%or adding 
and
colour jitter to existing data, where  
%DA has been shown  
the performance gains heavily relies on the choice of augmentation operations and parameters \cite{alexey2015discriminative}. 

%\par
To mitigate this reliance, 
recent efforts have focused on learning optimal augmentation operations for a given task and dataset \cite{cubuk2019autoaugment,zhang2020adversarial,ho2019population,lim2019fast}. However, transformations learned from these methods are typically still limited to a predefined set of simple operations such as rotation, translation, and scaling. 
In the meantime, another direction of research has emerged that provides an alternative way of learning more expressive augmentations based on deformation-based transformations commonly used in image registration \cite{hauberg2016dreaming,Zhao_2019_CVPR,10.1007/978-3-030-59716-0_31}. 
Instead of pre-specifying a list of operations such as rotation and scaling \cite{cubuk2019autoaugment}, 
these deformation-based transformations 
can describe more general spatial transformations. Moreover, they are perfectly suited for modelling an object's shape changes \cite{Zhao_2019_CVPR} that are crucial for image segmentation tasks. It thus provides an excellent candidate for learning shape variations of an object from the data, and via which to enable shape-based augmentations for medical image segmentation tasks. 
\cite{Zhao_2019_CVPR,10.1007/978-3-030-59716-0_31}.

However, to date, 
all existing approaches to learning deformable registration-based DA 
assume a perfect alignment of image pairs to learn the transformations. 
In other words, the deformation-based transformations are learned globally for the image.
This assumption is restrictive and associated with several challenges. First, the learning of a global image-level transformation requires image alignment that may be non-trivial in many scenarios, such as 
the alignment of tumours that can appear at different locations of an image, or alignment of images from different modalities. The learning of transformations itself is also complicated by the presence of other objects in the image and is best suited when the object of interest is always in the same (and often centre) location in all the images, \textit{i.e.,} images are globally aligned \textit{a priori} \cite{Zhao_2019_CVPR}. 
Second, the application of the learned global transformations for DA is also restricted to images similar (and aligned) to those in training.
 It thus will be challenging to transfer the learned shape variations to even the same objects across different locations, orientations, or sizes in the image, let alone transferring across dataset (\textit{e.g.}, to transfer the learned shape variations of an organ from one image modality to another). 

%\par
Intuitively, object-centric transformations and augmentations have the potential to overcome the challenges associated with global image-level transformations. 
Recently, an object-centric augmentation method termed as TumorCP \cite{10.1007/978-3-030-87193-2_55} showed that a simple object-level augmentation, via copy-pasting a tumour from one location to another, can yield impressive performance gains. 
However, the diversity of samples generated by TumorCP is limited to pasting tumours on different backgrounds with random distortions without further learned shape-based augmentation.

Similarly, other existing works on object-level augmentation of lesions have mostly focused on 
 position, orientation, and random transformations of the lesion on different backgrounds \cite{10.1007/978-3-030-87193-2_19,10.1007/978-3-031-16440-8_65}. 
 To date, no existing works
have considered shape-based 
object-centric augmentations. 
Enriching object-centric DA  with learned shape variations -- a factor critical to object segmentation -- can result in  more diverse samples and thereby improve DNN training for medical image segmentation. 

\begin{figure}[!tb]
    \begin{center}
        \includegraphics[width=.6\textwidth,scale=0.3]{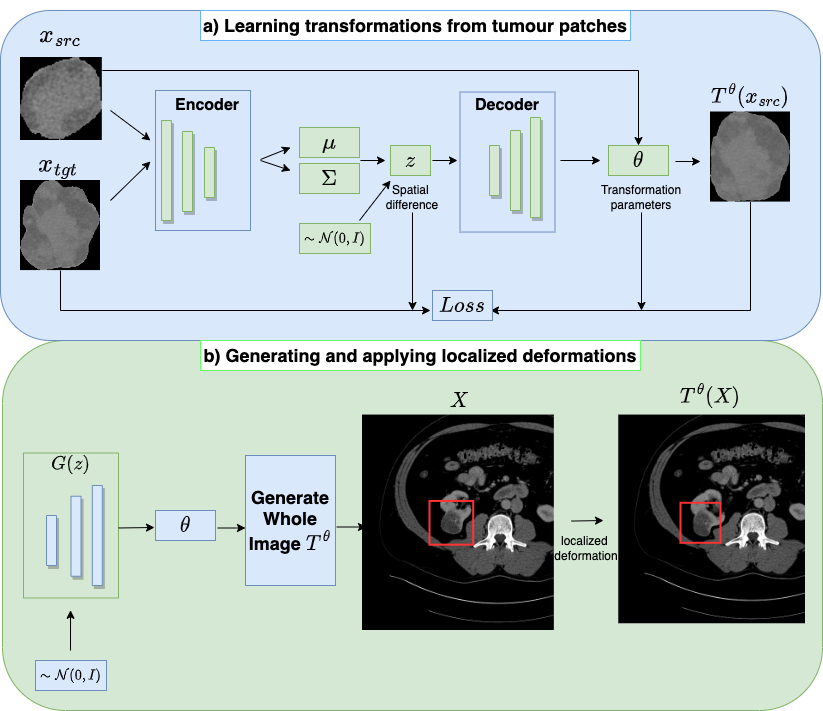}
        \caption{Overview of the presented approach. a) Learning a generative model describing object-centric shape variations as diffeomorphic transformations. b) Sampling transformations from the learned generative model and deforming an object in place (highlighted with a red square) without distorting the surrounding content in the image. 
        }
        \label{fig:overview}
    \end{center}
\vspace{-0.8cm}
\end{figure}

In this paper, we present a novel approach for learning and transferring object-centric deformations for DA in medical image segmentation tasks. As illustrated in Fig.~\ref{fig:overview}, 
this is achieved with two key elements: 
%1) 
\begin{itemize}

     \item A generative model of object-centric deformations -- constrained to C1 diffeomorphism for better DNN training -- to describe 
shape variability learned from paired patches of objects of interest. 
This allows the learning to focus on the shape variations of an object regardless of its positions and sizes in the image, thus bypassing the requirement for image alignment. 

     \item An online augmentation strategy to 
     sample transformations from the generative model and to
     %Secondly, we use transformations from the generative model to
     augment the objects of interest in place without distorting the surrounding content in the image. This allows us to add shape diversity to the objects of interest in an image regardless of their positions or sizes, eventually facilitating transferring the learned variations across datasets.

\end{itemize}

We demonstrated the effectiveness of the presented object-centric diffeomorphic augmentation in kidney tumour segmentation, 
including using shape variations of kidney tumours 
learned from the same dataset (KiTS \cite{https://doi.org/10.48550/arxiv.1904.00445}), as well as transferring those learned from a larger liver tumour dataset (LiTS \cite{BILIC2023102680}). Experimental results showed that it
can enrich the augmentation diversity of other techniques such as TumorCP \cite{10.1007/978-3-030-87193-2_55}, 
and improve
kidney tumour segmentation \cite{https://doi.org/10.48550/arxiv.1904.00445} using shape variations learned either within or outside the same training data. % used for the segmentation task.     

\section{Methods}

We focus on DA for tumour segmentation because tumours can occur at different locations of an organ with substantially different orientations and sizes. It thus presents a challenging scenario where global image-level deformable transformations cannot apply.
mentation approach comprises as outlined in 
Below we describe the two key methodological elements.

\subsection{Object-Centric Diffeomorphism as a Generative Model}

The goal of this element is to learn 
to generate diffeomorphic transformation parameters $\theta$ that describe shape variations -- in the form of deformable transformations $T^\theta$ -- 
that are present  
within training instances of tumour $x$. 
To realize this, we train a generative model $G(.)$ for $\theta$ such that, 
when given %the spatial difference between 
two instances of tumours ($x_{src}, x_{tgt}$), 
it is asked to generate $\theta$ from the encoded latent representations $z$ in order to deform $x_{src}$ through $T^{\theta}(x_{src})$ to $x_{tgt}$. %$ \simeq T^{\theta}(x_{src})$.

\subsubsection{Transformations:}
In order to model shape deformations between $x_{src}$ and $x_{tgt}$, we need highly expressive transformations to capture rich shape variations in tumour pairs. We assume a spatial transformation $T^\theta$ in the form of pixel-wise displacement field $u$ as $T^\theta(x) = x + u$. Inspired from \cite{8578561,hauberg2016dreaming}, we turn to $C^1$ diffeomorphisms to model our transformations. $C^1$ diffeomorphisms are smooth and invertible transformations that preserve differentiability up to the first derivative, making them a suitable choice to be embedded in a neural network for gradient-based optimization \cite{8578561}. However, the set of all diffeomorphisms is an infinitely large Lie group. To overcome this issue, we focus on a specific finite-dimensional subset of the Lie group that is large enough to capture the relevant variations in the tumours. For this, we make use of continuous piecewise-affine-based (CPAB) transformation based on the integration of CPA velocity field $v^\theta$ proposed in \cite{7410690}. Let $\Omega$ $\subset$ $R^{2}$ denote the tumour domain and let $P$ be triangular tesselation of $\Omega$ \cite{hauberg2016dreaming}. A velocity field that maps points from $\Omega$ to $R^2$ is said to be piecewise affine (PA) if it is affine when restricted to each triangle of $P$. The set $V$ of $v^\theta$ which are zero on the boundary of $\Omega$ can be shown to be finite-dimensional linear space \cite{7410690}. The dimensionality $d$ of $V$ is a result of how fine $P$ is tessellated. It can be shown that $V$ is parameterized by $\theta$, \textit{i.e.,} any instance of $V$ is a linear combination of $d$ orthonormal CPA fields with weights $\theta$ \cite{7410690}. A spatial transformation $T^\theta$ can be derived by integrating a velocity field $v^\theta$ \cite{7410690} as: %\ref{eq:1},  %for integrating $v^\theta$ 

\begin{equation} \label{eq:1}
u^{\theta}(x,t) = 
x + \int_{0}^{t} v^{\theta} (u^{\theta}(x,t)) dt  
\end{equation}
where the integration can be done via a specialized solver \cite{7410690}. The solver chosen produces faster and more accurate results than a generic ODE solver. Specifically, the cost for this solver is $O(C1) + O(C2$ x Number of integration steps), where $C1$ is matrix exponential for the number of cells an image is divided into and $C2$ is the dimensionality of an image. The transformations $T^\theta$ thus can be described by a generative model of $\theta$. We also experimented with squaring and scaling layers for integration but that resulted in texture loss when learning transformations. 

\subsubsection{Generative Modeling:} The data generation process can be described as:
%as illustrated in Fig.~\ref{fig:overview}(a) is: 
\begin{eqnarray}
\label{gen_model}
& p(z) \sim 
\mathcal{N}(0,I), \quad
\theta \sim  p_\phi(\theta|z), \quad
x_{tgt} \sim
p(x_{tgt}|\theta, x_{src})\\%\quad
& p(x_{tgt}, z | x_{src})
= p(z) \int_\theta p(x_{tgt}|\theta, x_{src})p_\phi(\theta|z)d\theta = p(z)p_\phi(x_{tgt}|z,x_{src}) 
\end{eqnarray} 
where $z$ is the latent variable assumed to follow an isotropic Gaussian prior, 
$p_\phi(\theta|z)$ is modeled by a neural network parameterized by $\phi$, and $p(x_{tgt}|\theta, x_{src})$ 
follows the deformable transformation as described in Equation
\eqref{eq:1}.

We 
define variational approximations of the posterior density 
%$p(z|x_{src},x_{tgt})$ 
as $q_\psi(z|x_{src},x_{tgt})$, modeled by a convolutional neural network that expects two inputs $x_{src}$ and $x_{tgt}$. Passing a tuple of $x_{src}$ and $x_{tgt}$ as the input helps the latent representations to learn the spatial difference between two tumour samples. 
Alternatively, the generative model as described can be considered as a conditional model where both the generative and inference model is conditioned on the source tumour sample $x_{src}$.

\subsubsection{Variational Inference:} 

The parameters $\psi$ and $\phi$ are optimized by the modified evidence lower bound (ELBO) of the log-likelihood $\log p(x_{tgt}|x_{src})$:

\begin{equation}
\label{eq:var_infr}
\begin{split}{}
\log p(x_{tgt}|x_{src}) \ge \mathcal{L}_{ELBO} = 
    E_{q_\psi(z|x_{src}, x_{tgt})} p_\phi(x_{tgt}|z,x_{src}) \\
    -  \beta {D}_{KL}(q_\psi(z|x_{src}, x_{tgt}) ||p(z))
\end{split}    
\end{equation} 

where the first term in the ELBO takes the form of similarity loss: $L_2$ norm on the difference between $x_{tgt}$ and $\hat{x}_{src} = T^\theta(x_{src})$ synthesized using the $\theta$ from $G(z)$. 

The second KL term constrains our approximated posterior $q_\psi(z|x_{src}, x_{tgt})$ to be closer to the isotropic Gaussian prior $p(z)$, and its contribution to the overall loss is scaled by 
the hyperparameter $\beta$. 
To further ensure that $\hat{x}_{src}$ looks realistic, 
we discourage $G(z)$ from generating overly-expressive transformations 
%that can result in unrealistic-looking transformed tumours 
by adding a regularization term over the $L_2$ norm of the displacement field $u$ with a tunable hyperparameter $\lambda_{reg}$.
The final objective function becomes: 
\begin{equation}
\label{eq:loss_fn}
   \mathcal{L} = \mathcal{L}_{ELBO} + \lambda_{reg}*\| u \|_2
\end{equation}

\vspace{-.3cm}
\subsubsection{Object-Centric Learning:}
To learn object-centric spatial transformations,
$x_{src}$ and $x_{tgt}$ are in the forms of 
image patches that solely contain tumours. Given an image and its corresponding tumour segmentation mask ($X,Y$), we first extract a bounding box around the tumour by applying skimage.measure.regionprops from the scikit-image package to $Y$. We then use this bounding box to carve out the tumour $x$ from the image $X$, masking out all the regions within the bounding box that do not belong to the tumour. All the tumour patches are then resized to the same scale, such that tumours of different sizes can be described by the same tesselation resolution.
%^in order to only focus shape variations among them. 
When pairing tumour patches, we pair each tumour with its $K$ nearest neighbour tumours based on their Euclidean distance -- this again avoids learning overly expressive transformation when attempting to deform between significantly different tumour shapes.

\subsection{Online Augmentations with Generative Models}

The goal of this element is to sample random object-centric transformations of $T^\theta$ from $G(z)$, to generate diverse augmentations of different instances of tumours in place. However, if we only transform the tumour and keep the rest of the image identical, the transformed tumour may appear unrealistic and out of place. To ensure that the entire transformed image appears smooth, we use a hybrid strategy to construct a deformation field for the entire image $X$ that combines tumour-specific deformations with an identity transform for the rest of the image. Specifically, we fill a small region around the tumour with displacements of diminishing magnitudes, achieved by propagating the deformations from the boundaries of the deformation fields from $G(z)$ to their neighbours with reduced magnitudes. Repeating this process ensures that the change at the boundaries is smooth and that the transformed region appears naturally as part of the image. 

\section{Experiments and Results}
We used two publicly available datasets, LiTS \cite{BILIC2023102680} and KiTS \cite{https://doi.org/10.48550/arxiv.1904.00445}, for our experiments. LiTS \cite{https://doi.org/10.48550/arxiv.1904.00445} contains liver and liver tumour segmentation masks for 200 scans in total, 130 train and 70 test. Similarly, KiTS \cite{BILIC2023102680} has kidney and kidney tumour segmentation masks for 300 scans, 168 train, 42 validation, and 90 test. We trained our generative model $G(z)$ on KiTS and LiTS separately to learn spatial variations in tumour shapes. We then used either of the learned transformation to augment kidney tumor segmentation tasks on subsets of KiTS data with varying sizes. Code link: https://github.com/nileshkumar0726/Learning\_Transformations

\subsection{Generative Model Implementation, Training, and Evaluation}

\begin{figure}[!tb]
    \centering
    \includegraphics[height=0.55\textwidth,scale=0.4]{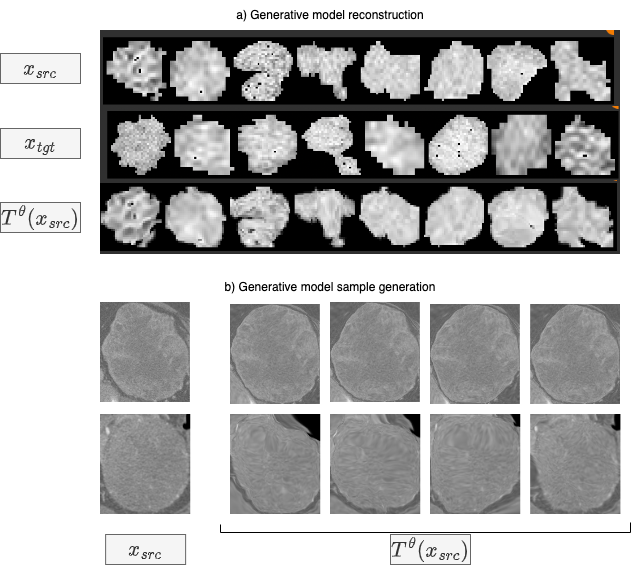}
    \caption{Visual examples of the generative model in (a) reconstructing $x_{tgt}$ given pairs of $x_{scr}$ and $x_{tgt}$, and (b) generating deformed samples given a single $x_{src}$.}
    \label{fig:recons}
\vspace{-0.5cm}
\end{figure}

\subsubsection{Data:} We prepared data for the generative model $G(z)$ training by first carving out tumour regions from individual slices of 3D scans using tumour segmentation masks. All tumour patches were resized to 30$\times$30, and each was paired with eight of its closest neighbours. We trained $G(z)$ with different sizes of data for individual experiments presented in Table ~\ref{seg_results}, ranging from using 11000 pairs from LiTS to only 3000 pairs when using only 25\% samples from KiTS. 

\subsubsection{Model:} The encoder of $G(z)$ consisted of five convolutional layers and three fully connected layers, with a latent dimension of 12 for $z$. The decoder consisted of five fully connected layers to output the parameters $\theta$ for $T^\theta$. We trained the $G(z)$ for a total of 400 epochs and a batch size of 16. We also implemented early stopping if the validation loss does not improve for 20 epochs. We used Adam optimizer \cite{https://doi.org/10.48550/arxiv.1412.6980} with a learning rate of 1e-4. 

We trained separate $G(z)$'s from KiTS and LiTS, respectively. We set $\beta$ = 0.001 for both models but needed a high $\lambda_{reg}$ of 0.009 for the LiTS model compared to 0.004 for KiTS model. The tumours in the LiTS have higher intensity differences, which may explain why a higher value of $\lambda_{reg}$ was needed to ensure that transformed tumours did not become unrealistic.

\subsubsection{Results:} 
We evaluated $G(z)$ with two criteria. First, the model needs to be able to reconstruct $x_{tgt}$ by generating $\theta$ to transform $x_{src}$. Second, the model needs to be able to generate diverse transformed tumour samples for a given tumour sample. Fig.~\ref{fig:recons} presents visual examples of the reconstruction and generation results achieved by $G(z)$.
%qualitative results for both these criteria in . 
It can be observed that the reconstruction is successful in most cases, except when $x_{src}$ and $x_{tgt}$ were too different. %, the transformation stops somewhere in the middle. 
This was necessary to ensure that $T^\theta(x_{src})$ did not produce unrealistic examples. 
The averaged L2-loss of transformed $\hat{x}_{src}$ was 1.23 
%After optimization, we end up with an average $L_2$ error of 1.23 
on the validation pairs.
%reduced from 1.7 at the beginning. 
We also visually inspected validation samples after training to make sure that the deformed tumours were similar to the original tumours in appearance.   
The generated examples of tumours from a single source, as shown in Fig.~\ref{fig:recons}(b), demonstrated that the generations were diverse yet realistic.

\subsection{Deformation-Based DA for Kidney Tumour Segmentation}

\begin{table}[t]
\begin{center}
\resizebox{\linewidth}{!}{ 
\begin{tabular}{ |c|c|c|c| } 
\hline
\bf{\% data for training} & \bf{Augmentations} & \bf{Mean Dice (std)}* $\uparrow{}$ \\
\hline
\multirow{4}{*}{25\%} & Baseline & 0.467 (0.014) \\ 
& Random Wrapping & 0.535 (0.003) \\
& TumorCP & 0.568 (0.014) \\ 
& Diffeo (within-data | cross-data) & 0.497 (0.006) | 0.505 (0.002) \\ 
%& Diffeo (LiTS) & 0.5051 (0.002) \\ 
& TumorCP + Diffeo (within-data | cross-data) & \textbf{0.581 (0.012)} | 0.576 (0.015) \\ 
%& TumorCP + Diffeo (LiTS) & 0.5766 (0.015) \\  
\hline
\multirow{4}{*}{50\%} & Baseline & 0.608 (0.017) \\
& Random Wrapping & 0.6675 (0.0091) \\
& TumorCP & 0.669 (0.011) \\ 
& Diffeo (within-data | cross-data) & 0.640 (0.002) | 0.639 (0.014) &  \\ 
%& Diffeo (LiTS) & 0.6391 (0.014)  \\ 
& TumorCP + Diffeo (within-data | cross-data) & 0.689 (0.013) | \textbf{0.702 (0.016)} \\ 
%& TumorCP + Diffeo (LiTS) & \textbf{0.702 (0.016)} \\
\hline
\multirow{4}{*}{75\%} & Baseline & 0.656 (0.027) \\ 
& Random Wrapping & 0.6774 (0.0036) \\
& TumorCP & 0.690 (0.003 \\ 
& Diffeo (within-data | cross-data) & 0.662 (0.006) | 0.655 (0.001) \\ 
%& Diffeo (LiTS) & 0.6556 (0.0011) \\ 
& TumorCP + Diffeo (within-data | cross-data) & 0.707 (0.001) | \textbf {0.718 (0.007)} \\ 
%& TumorCP + Diffeo (LiTS) & \textbf {0.718 (0.0076)} \\
\hline
\multirow{4}{*}{100\%} & Baseline & 0.680 (0.025) \\
& Random Wrapping & 0.6698 (0.016) \\
& TumorCP & 0.702 (0.005) \\ 
& Diffeo (within-data | cross-data) & 0.687 (0.014) | 0.688 (0.028) \\ 
%& Diffeo (LiTS) & 0.6885 (0.0284) \\ 
& TumorCP + Diffeo (within-data | cross-data) & 0.709 (0.004) | \textbf{0.713 (0.019)} \\ 
%& TumorCP + Diffeo (LiTS) & \textbf{0.713 (0.019)} \\
\hline
\end{tabular}
}
\end{center}
\caption{KiTS segmentation results in terms of DICE score. The baseline model already includes standard data augmentations. Within-data augmentations used transformations learned from KiTS using the same \% of training data for segmentation tasks. Cross-data augmentations used transformations learned from LiTS. TumorCP was also always performed within data.}
\label{seg_results}
\vspace{-0.8cm}
\end{table}

\subsubsection{Data:} 
We then used $G(z)$ to generate deformation-based augmentations to increase the size and diversity of training samples for kidney tumour segmentation on KiTS. 
To assess the effect of augmentations on 
different sizes of labelled data, 
we considered training using 25\%, 50\%, 75\%, and 100\% of the KiTS training set. We considered two DA scenarios: 
augment with transformations learned from KiTS (within-data augmentation) versus from LiTS (cross-data augmentation).

\subsubsection{Models:} For the base segmentation network, we adopted nnU-net \cite{isensee2021nnu} as it contains state of the art (SOTA) pipeline for medical image segmentation on most datasets. To make the segmentation pipeline compatible with $G(z)$, we used the 2D segmentation module of nnU-net. 
For baselines, 
 we considered 1) default augmentations such as rotation, scaling, and random crop in nnU-net as well as 2)
 TumorCP, all modified for 2D segmentation.
Note that our goal is not to achieve SOTA results on 
KiTS, but to test the relative efficacy of the presented DA 
strategies in comparison with existing object-centric DA methods.

\subsubsection{Results:}
We use Sørensen-Dice Coefficient (Dice) to measure segmentation network performance. Dice measures the overlap between prediction and ground truth. As summarized in Table~\ref{seg_results}, 
when combined with TumorCP, 
the presented augmentations 
were able to generate statistically significant (paired \textit{t}-test, $ p \le 0.05 $) improvements in all cases 
compared to TumorCP alone. This demonstrated the benefit of enriching simple copy-and-paste DA with shape variations. 
Interestingly, 
cross-data transferring of the learned augmentations (from LiTS) outperformed the within-data augmentation in the majority of the cases. Which we believe is because of two factors. Firstly, learning of the within-data augmentations is limited to the percentage of the training set used for segmentation. The number of objects to learn transformations from is thus greater in cross-data augmentation settings. Secondly, the transformations present in cross-data are completely unseen in the segmentation training network which helps in generating more diverse samples. Note that, as the transformations are learned as variations in object shapes, they can be transferred easily across datasets

%variations across datasets. 
Surprisingly, 
the improvements achieved by the presented augmentation strategy were the most prominent when the segmentation was trained on 50\% and 75\% of the KiTS training set. 
This is contrary to the expectation that DA would be most beneficial when the labelled training set is small.  
This may be because smaller sample sizes do not provide sufficient initial tumor samples for shape transformations. This may also explain why the combination of TumorCP boosted the performance of our augmentation strategy, as 
the oversampling nature of TumorCP provided more tumour samples for 
the presented strategy to transform to further enrich the training set. It is also worth noting that in contrast to prior literature, random wrapping of objects does not come close to the learned augmentations. We speculate that while unrealistic transformations work for whole images, they may be problematic when only augmenting specific local objects in an image.           
\section{Discussion and Conclusions}

In this work, we presented a novel diffeomorphism-based object-centric augmentation that can be learned and used to augment the objects of interest regardless of their position and size in an image. As demonstrated by the experimental results, this allows us to not only introduce new variations to unfixed objects like tumours in an image but also transfer the knowledge of shape variations across datasets. 
An immediate next step will be to extend the presented approach to learn and transfer
3D transformations for 3D segmentation tasks, and to enrich the shape-based transformation with appearance-based transformations. In the long term, it would be interesting to explore ways to transfer knowledge about more general forms of variations across datasets.

\section{Acknowledgments}

This work is supported by the National Institute of Nursing Research (NINR) of the National Institutes of Health (NIH) under Award Number R01NR018301.


\bibliographystyle{splncs04}
\bibliography{ref}

\begin{thebibliography}{10}
\providecommand{\url}[1]{\texttt{#1}}
\providecommand{\urlprefix}{URL }
\providecommand{\doi}[1]{https://doi.org/#1}

\bibitem{alexey2015discriminative}
Alexey, D., Fischer, P., Tobias, J., Springenberg, M.R., Brox, T.:
  Discriminative unsupervised feature learning with exemplar convolutional
  neural networks. IEEE Trans. Pattern Analysis and Machine Intelligence
  \textbf{99} (2015)

\bibitem{BILIC2023102680}
Bilic, P., Christ, P., Li, H.B., Vorontsov, E., Ben-Cohen, A., Kaissis, G.,
  Szeskin, A., Jacobs, C., Mamani, G.E.H., Chartrand, G., Lohöfer, F., Holch,
  J.W., Sommer, W., Hofmann, F., Hostettler, A., Lev-Cohain, N., Drozdzal, M.,
  Amitai, M.M., Vivanti, R., Sosna, J., Ezhov, I., Sekuboyina, A., Navarro, F.,
  Kofler, F., Paetzold, J.C., Shit, S., Hu, X., Lipková, J., Rempfler, M.,
  Piraud, M., Kirschke, J., Wiestler, B., Zhang, Z., Hülsemeyer, C., Beetz,
  M., Ettlinger, F., Antonelli, M., Bae, W., Bellver, M., Bi, L., Chen, H.,
  Chlebus, G., Dam, E.B., Dou, Q., Fu, C.W., Georgescu, B., i~Nieto, X.G.,
  Gruen, F., Han, X., Heng, P.A., Hesser, J., Moltz, J.H., Igel, C., Isensee,
  F., Jäger, P., Jia, F., Kaluva, K.C., Khened, M., Kim, I., Kim, J.H., Kim,
  S., Kohl, S., Konopczynski, T., Kori, A., Krishnamurthi, G., Li, F., Li, H.,
  Li, J., Li, X., Lowengrub, J., Ma, J., Maier-Hein, K., Maninis, K.K., Meine,
  H., Merhof, D., Pai, A., Perslev, M., Petersen, J., Pont-Tuset, J., Qi, J.,
  Qi, X., Rippel, O., Roth, K., Sarasua, I., Schenk, A., Shen, Z., Torres, J.,
  Wachinger, C., Wang, C., Weninger, L., Wu, J., Xu, D., Yang, X., Yu, S.C.H.,
  Yuan, Y., Yue, M., Zhang, L., Cardoso, J., Bakas, S., Braren, R., Heinemann,
  V., Pal, C., Tang, A., Kadoury, S., Soler, L., {van Ginneken}, B., Greenspan,
  H., Joskowicz, L., Menze, B.: The liver tumor segmentation benchmark (lits).
  Medical Image Analysis  \textbf{84},  102680 (2023).
  \doi{https://doi.org/10.1016/j.media.2022.102680},
  \url{https://www.sciencedirect.com/science/article/pii/S1361841522003085}

\bibitem{cubuk2019autoaugment}
Cubuk, E.D., Zoph, B., Mane, D., Vasudevan, V., Le, Q.V.: Autoaugment: Learning
  augmentation strategies from data. In: Proceedings of the IEEE conference on
  computer vision and pattern recognition. pp. 113--123 (2019)

\bibitem{8578561}
Detlefsen, N.S., Freifeld, O., Hauberg, S.: Deep diffeomorphic transformer
  networks. In: 2018 IEEE/CVF Conference on Computer Vision and Pattern
  Recognition. pp. 4403--4412 (2018). \doi{10.1109/CVPR.2018.00463}

\bibitem{7410690}
Freifeld, O., Hauberg, S., Batmanghelich, K., Fisher, J.W.: Highly-expressive
  spaces of well-behaved transformations: Keeping it simple. In: 2015 IEEE
  International Conference on Computer Vision (ICCV). pp. 2911--2919 (2015).
  \doi{10.1109/ICCV.2015.333}

\bibitem{hauberg2016dreaming}
Hauberg, S., Freifeld, O., Larsen, A.B.L., Fisher, J., Hansen, L.: Dreaming
  more data: Class-dependent distributions over diffeomorphisms for learned
  data augmentation. In: Artificial Intelligence and Statistics. pp. 342--350
  (2016)

\bibitem{https://doi.org/10.48550/arxiv.1904.00445}
Heller, N., Sathianathen, N., Kalapara, A., Walczak, E., Moore, K., Kaluzniak,
  H., Rosenberg, J., Blake, P., Rengel, Z., Oestreich, M., Dean, J., Tradewell,
  M., Shah, A., Tejpaul, R., Edgerton, Z., Peterson, M., Raza, S., Regmi, S.,
  Papanikolopoulos, N., Weight, C.: The kits19 challenge data: 300 kidney tumor
  cases with clinical context, ct semantic segmentations, and surgical outcomes
  (2019). \doi{10.48550/ARXIV.1904.00445},
  \url{https://arxiv.org/abs/1904.00445}

\bibitem{ho2019population}
Ho, D., Liang, E., Stoica, I., Abbeel, P., Chen, X.: Population based
  augmentation: Efficient learning of augmentation policy schedules. arXiv
  preprint arXiv:1905.05393  (2019)

\bibitem{isensee2021nnu}
Isensee, F., Jaeger, P.F., Kohl, S.A., Petersen, J., Maier-Hein, K.H.: nnu-net:
  a self-configuring method for deep learning-based biomedical image
  segmentation. Nature methods  \textbf{18}(2),  203--211 (2021)

\bibitem{https://doi.org/10.48550/arxiv.1412.6980}
Kingma, D.P., Ba, J.: Adam: A method for stochastic optimization (2014).
  \doi{10.48550/ARXIV.1412.6980}, \url{https://arxiv.org/abs/1412.6980}

\bibitem{lim2019fast}
Lim, S., Kim, I., Kim, T., Kim, C., Kim, S.: Fast autoaugment. In: Advances in
  Neural Information Processing Systems. pp. 6662--6672 (2019)

\bibitem{10.1007/978-3-030-59716-0_31}
Shen, Z., Xu, Z., Olut, S., Niethammer, M.: Anatomical data augmentation via
  fluid-based image registration. In: Martel, A.L., Abolmaesumi, P., Stoyanov,
  D., Mateus, D., Zuluaga, M.A., Zhou, S.K., Racoceanu, D., Joskowicz, L.
  (eds.) Medical Image Computing and Computer Assisted Intervention -- MICCAI
  2020. pp. 318--328. Springer International Publishing, Cham (2020)

\bibitem{10.1007/978-3-030-87193-2_55}
Yang, J., Zhang, Y., Liang, Y., Zhang, Y., He, L., He, Z.: Tumorcp: A simple
  but effective object-level data augmentation for tumor segmentation. In:
  de~Bruijne, M., Cattin, P.C., Cotin, S., Padoy, N., Speidel, S., Zheng, Y.,
  Essert, C. (eds.) Medical Image Computing and Computer Assisted Intervention
  -- MICCAI 2021. pp. 579--588. Springer International Publishing, Cham (2021)

\bibitem{10.1007/978-3-030-87193-2_19}
Zhang, X., Liu, C., Ou, N., Zeng, X., Xiong, X., Yu, Y., Liu, Z., Ye, C.:
  Carvemix: A simple data augmentation method for brain lesion segmentation.
  In: de~Bruijne, M., Cattin, P.C., Cotin, S., Padoy, N., Speidel, S., Zheng,
  Y., Essert, C. (eds.) Medical Image Computing and Computer Assisted
  Intervention -- MICCAI 2021. pp. 196--205. Springer International Publishing,
  Cham (2021)

\bibitem{zhang2020adversarial}
Zhang, X., Wang, Q., Zhang, J., Zhong, Z.: Adversarial autoaugment. In:
  International Conference on Learning Representations (2020),
  \url{https://openreview.net/forum?id=ByxdUySKvS}

\bibitem{Zhao_2019_CVPR}
Zhao, A., Balakrishnan, G., Durand, F., Guttag, J.V., Dalca, A.V.: Data
  augmentation using learned transformations for one-shot medical image
  segmentation. In: Proceedings of the IEEE/CVF Conference on Computer Vision
  and Pattern Recognition (CVPR) (June 2019)

\bibitem{10.1007/978-3-031-16440-8_65}
Zhu, Q., Wang, Y., Yin, L., Yang, J., Liao, F., Li, S.: Selfmix: A
  self-adaptive data augmentation method for lesion segmentation. In: Wang,
  L., Dou, Q., Fletcher, P.T., Speidel, S., Li, S. (eds.) Medical Image
  Computing and Computer Assisted Intervention -- MICCAI 2022. pp. 683--692.
  Springer Nature Switzerland, Cham (2022)

\end{thebibliography}
\end{document}